\documentclass[letterpaper, 10 pt, conference]{ieeeconf}  % Comment this line out if you need a4paper

\IEEEoverridecommandlockouts                              % This command is only needed if 
\usepackage{algorithm}
\usepackage[noend]{algpseudocode}
\usepackage{subcaption}
\usepackage{mwe}
\usepackage{breqn}

\usepackage{amsthm,amsmath}
\usepackage{amssymb}
\let\labelindent\relax
\usepackage[shortlabels]{enumitem}
\theoremstyle{definition}

\usepackage{dsfont}
\usepackage{tikz-network}
\newtheorem{problem}{Problem}

\newtheorem{remark}{Remark}

\newtheorem{definition}{Definition}
\usepackage{balance}

\newcommand{\mc}[1]{\mathcal{#1}}
\DeclareMathOperator*{\argmin}{\arg\!\min}
\DeclareMathOperator*{\argmax}{\arg\!\max}

\newcommand{\mT}{\mathcal{T}}

\usepackage{xcolor}
\usepackage{hyperref} 
\makeatletter
\newcommand\fs@betterruled{%
  \def\@fs@cfont{\bfseries}\let\@fs@capt\floatc@ruled
  \def\@fs@pre{\vspace*{5pt}\hrule height.8pt depth0pt \kern2pt}%
  \def\@fs@post{\kern2pt\hrule\relax}%
  \def\@fs@mid{\kern2pt\hrule\kern2pt}%
  \let\@fs@iftopcapt\iftrue}
\floatstyle{betterruled}
\restylefloat{algorithm}
\makeatother

\title{\LARGE \bf
Synthesizing Reactive Test Environments for Autonomous Systems:  
Testing Reach-Avoid Specifications with Multi-Commodity Flows}

\author{Apurva Badithela*$^{1}$, Josefine B. Graebener*$^{2}$, Wyatt Ubellacker$^{1}$, Eric V. Mazumdar$^{1}$,\\ Aaron D. Ames$^{1}$, Richard M. Murray$^{1}$% <-this % stops a space
\thanks{* The authors contributed equally. Corresponding authors: A. Badithela, J.B. Graebener \texttt{\{apurva,jgraeben\}@caltech.edu}}
\thanks{We acknowledge funding from AFOSR Test and Evaluation Program, grant FA9550-19-1-0302,  National Science Foundation award CNS-1932091, and Dow (\#227027AT).}
\thanks{$^{1}$Department of Computing and Mathematical Sciences, California Institute of Technology, Pasadena, CA 91125, USA}
\thanks{$^{2}$Graduate Aerospace Laboratories, California Institute of Technology, Pasadena, CA 91125, USA}
}

\begin{document}
\maketitle
\thispagestyle{empty}
\pagestyle{empty}

%%%%%%%%%%%%%%%%%%%%%%%%%%%%%%%%%%%%%%%%%%%%%%%%%%%%%%%%%%%%%%%%%%%%%%%%%%%%%%%%
\begin{abstract}
We study automated test generation for verifying discrete decision-making modules in autonomous systems. We utilize linear temporal logic to encode the requirements on the system under test in the system specification and the behavior that we want to observe during the test is given as the test specification which is unknown to the system. First, we use the specifications and their corresponding non-deterministic Büchi automata to generate the specification product automaton.
Second, a virtual product graph representing the high-level interaction between the system and the test environment is constructed modeling the product automaton encoding the system, the test environment, and specifications. 
The main result of this paper is an optimization problem, framed as a multi-commodity network flow problem, that solves for constraints on the virtual product graph which can then be projected to the test environment. Therefore, the result of the optimization problem is reactive test synthesis that ensures that the system meets the test specifications along with satisfying the system specifications.
% We provide an algorithm that finds the projection of the acceptance conditions of the system and test specifications on the game graph. 
% Finally, to ensure that the system meets the test specification in addition to satisfying the system specification, we present a framework to find constraints on the virtual product graph which will then be projected onto the test environment. 
% Specifically, we formulate this as a multi-commodity network flows problem, and present an optimization to solve for the constraints on the virtual product graph. 
This framework is illustrated in simulation on grid world examples, and demonstrated on hardware with the Unitree A1 quadruped, wherein dynamic locomotion behaviors are verified in the context of reactive test environments. 
% We conclude with future directions on applying these algorithms to constrain test environments in self-driving applications.
\end{abstract}
% \textcolor{blue}{Similar with few changes}
%%%%%%%%%%%%%%%%%%%%%%%%%%%%%%%%%%%%%%%%%%%%%%%%%%%%%%%%%%%%%%%%%%%%%%%%%%%%%%%%

\section{Introduction}
\label{sec:introduction}
\begin{figure}[]
    \centering
    \includegraphics[width = \columnwidth]{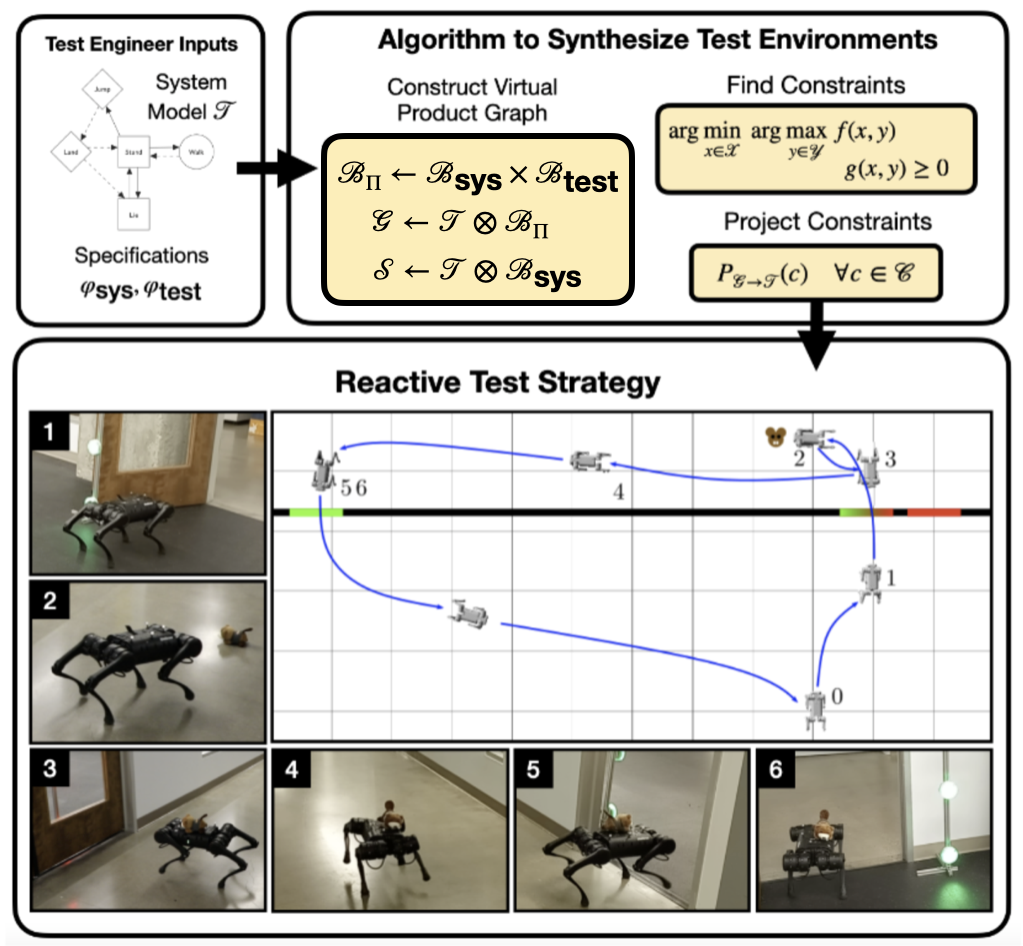}
    \caption{Overview of the test environment synthesis framework and the hardware demonstration.}
    \vspace{-5mm}
    \label{fig:overview}
\end{figure}
Operational testing of autonomous systems at various levels of abstraction---  from low-level continuous dynamics to high-level discrete decision-making--- is essential for verification and validation. 
In formal methods, testing refers to simulation-based falsification, where inputs to a model of the system are found which result in system violating its requirements~\cite{sankaranarayanan2012falsification,kapinski2016simulation,annpureddy2011s,chou2018using,dang2009coverage,hekmatnejad2020search,plaku2013falsification}. Falsification methods typically minimize a robustness metric associated with the formal specifications of the system to find inputs that result in falsifying traces~\cite{donze2010breach,fainekos2009robustness,dreossi2019verifai}.  However,  another  approach to testing is to have test engineers hand-design test scenarios as seen in the qualification tests of the DARPA Urban Challenge~\cite{DARPA_Urban_rules,DARPA_Urban}. 
In this work, we bridge these two approaches by leveraging test engineer expertise at the specification level and then automating the construction of the test environment for testing discrete, long-horizon decision-making in robotic systems. In the last decade, the control synthesis community has demonstrated the effectiveness of using temporal logic to specify formal requirements for robotic systems~\cite{kress2009temporal,wongpiromsarn2012receding,kloetzer2007temporal,lahijanian2015time}. Furthermore, we assume that via the use of rulebooks and industry standard manuals~\cite{censi2019liability,shalev2017formal,wongpiromsarn2021minimum}, a test engineer can provide these high-level descriptions on the desired test outcomes using temporal logic. 

Our notion of testing in this work is also complementary to falsification ---  we seek to construct a test environment to observe a desired high-level behavior, on which falsification can then be applied to determine the worst-case scenario which can provide confidence that the system will behave correctly in operational environments. Furthermore, falsification algorithms oftentimes search over continuous inputs to find a failure case~\cite{ernst2020arch,ghosh2018verifying,zutshi2014multiple,corso2021survey}. We envision that falsification could be used to refine the test environments generated by our approach. 

Our approach to test generation shares similarities with existing methods, but has key differences. Similar to~\cite{graebener2022towards}, we characterize the mission requirements on the system as a system specification, and characterize the desired behavior observed during the test via a test specification, which is unknown to the system. However, unlike~\cite{graebener2022towards}, we seek to construct the test environment, by constraining actions that the system can take, such that: a) the system can still satisfy its requirements, and b) the test specification is satisfied in a successful test execution (\ref{prob1}). Additionally, we seek to synthesize tests in which the system is not too restricted in its decision-making (\ref{prob2}).

% In this paper, we consider reach-avoid specifications in linear temporal logic (LTL). Using automata theory, we construct a virtual product graph on which a system trace satisfying the system and test specifications is equivalent to the trace reaching two different sets of acceptance states. Synthesizing a test by constraining system actions becomes equivalent to the problem of finding edge cuts on the virtual product graph such that a test execution that successfully reaches the system acceptance states also visits the tester acceptance states.
% To find edge cuts, we use multi-commodity network flows to model our problem. While a brute-force approach to solving this problem results in a combinatorial optimization problem, we present an efficiently solvable convex-concave min-max optimization-based relaxation that results in a constrained test. In particular, we relax the edge cuts to fractional values and present a bilevel optimization with a linear objective function and linear constraints. 
Building upon the results of~\cite{apurva2022mincons}, the key contributions of this paper are (i) framing the problem of synthesizing test environments for reach-avoid specifications in linear temporal logic (LTL) as a multi-commodity network flow problem, (ii) presenting an efficiently solvable convex-concave min-max optimization-based relaxation that results in a constrained test, (iii) demonstrating the approach by executing the resulting test strategy to reactively test dynamic locomotion behaviors of the Unitree A1 quadruped.
A key advantage of our method is that the synthesized test is reactive --- the constraints visible to the system under test are reactive to the system state and depend on the system's strategy, which is not known to the tester \emph{a priori}.

% \textcolor{blue}{Add more notes on related/previous work.}

% \begin{figure}
%     \centering
%     \includegraphics[width = \columnwidth]{figs/overview_v2.pdf}
%     \caption{Overview of the test environment synthesis framework and the hardware demonstration.}
%     \label{fig:overview}
% \end{figure}

\section{Background}
\label{sec:Prelims}
\vspace{-1mm}
\subsection{Temporal Logic, Transition Systems, and Automata}
\vspace{-1mm}
Linear temporal logic (LTL) can describe temporal properties on a trace of propositional formulas~\cite{baier2008principles}. The syntax of LTL is comprised of both logical (\(\wedge\) \emph{and}, \(\vee\) \emph{or}, and \(\neg\) \emph{negation}) and
temporal operators (\(\bigcirc\) \emph{next}, \(\square\) \emph{always}, \(\lozenge\) \emph{eventually}, and \(\mathcal{U}\) \emph{until}) operators. LTL can specify requirements on high-level decision-making in autonomous systems such as \emph{safety} \(\square(\varphi^s_{\text{sys}})\), \emph{progress} \(\lozenge(\varphi^p_{\text{sys}})\), and \emph{fairness} \(\square \lozenge (\varphi^f_{\text{sys}})\).

A nondeterministic B\"uchi automaton (NBA) \cite{Buchi1990} is a tuple \(\mc{B} = (Q, 2^{AP}, \delta, Q_0, F)\), where \(Q\) represents the states, \(AP\) is the set of atomic propositions, \(\delta\) represents the transition function, \(Q_0 \subseteq Q\) represents the initial states, and \(F\subseteq Q\) is the set of acceptance states. A transition system is a tuple \(\mT = (S, A, E, I, AP, L)\) where \(S\) is a set of states, \(A\) is the set of actions, \(E: S \times A \rightarrow S\) is the transition relation, \(I \subseteq S\) is the set of initial states, \(AP\) is the set of atomic propositions, and \(L:S\rightarrow 2^{AP}\) is a labeling function that indicates the set of atomic propositions that evaluate to \emph{true} at a particular state.

\begin{definition}[Product Automaton] A \emph{product automaton} is the synchronous product of a transition system \(\mc{T}=(S, A, E, I, AP, L)\) and a NBA \(\mc{B} = (Q, 2^{AP}, \delta, Q_0, F)\), is the tuple \(\mc{T} \otimes \mc{B} = (S', A, E', I', AP', L')\), where: 
\begin{itemize}
    \item \(S' = S\times Q\),
    \item \(\forall s,t \in S\), \(\forall q,p \in Q\) such that \(s\overset{a}{\rightarrow} t\) and \(\delta(q, L(t)) = p\), then, \((s,q)\overset{a}{\rightarrow}' (t,p)\), 
    \item \(I' = \{(s_0,q): s_0 \in I,\, \exists q_0\in Q_0 \text{ s.t. } q_0\overset{L(s_0)}{\rightarrow} q\}\),
    \item \(AP'=Q\), and
    \item \(L':S\times Q \rightarrow 2^Q\) such that \(L'((s,q)) = \{q\}\).
\end{itemize} 
\end{definition}

\begin{definition}[Asynchronous Product Automaton] An \emph{asynchronous product automaton} of finite state automata \(\mc{A}_1, \ldots, \mc{A}_n\) is the finite state automaton \(\mc{A}_{\Pi} = (S_{\Pi}, s_{0, \Pi}, L_{\Pi}, F_{\Pi}, T_{\Pi})\), where:
\begin{itemize}
    \item \(S_{\Pi} := S_1 \times \ldots \times S_n\), the Cartesian product of the states of the individual product automata, 
    \item \(s_{0, \Pi} = (s_{01}, \ldots, s_{0n})\), the \(n-\)tuple representing initial conditions,
    \item \(L_{\Pi} := L_1 \cup \ldots \cup L_n\), where \(L_i = 2^{AP_i}\) for all \(i \in \{1, \ldots, n\}\),
    \item \(T_{\Pi} := ((u_1, \ldots, u_n), l, (v_1, \ldots, v_n))\) such that \(\exists i\) and \(1 \leq i \leq n\), \((u_i, l, v_i) \in T_i\) and \(\forall j \neq i\), \(u_j = v_j\),
    \item \(F_{\Pi} := (s_1, \ldots, s_n) \in S\) such that \(\exists i\) such that \(s_i \in F_i\).
\end{itemize}
\end{definition}

\subsection{System and Test Environment}
\vspace{-1mm}
% \textcolor{blue}{Shorten.}
We utilize the notion of a system specification and a test specification, which represent requirements on the system under test and the test environment, respectively~\cite{graebener2022towards}. The system specification is assumed to be given, while the system does not necessarily have knowledge of the entire test specification.
In this work we will frame the system specification and the test specification as reach-avoid type specifications defined as
\begin{equation}
    \varphi_{\text{sys}} = \square\varphi^s_{\text{sys}} \land \lozenge\varphi^p_{\text{sys}},\: 
    \varphi_{\text{test}} = \square(\varphi^s_{\text{test}}) \land \bigwedge_i\lozenge(\varphi^p_{\text{test}})_i,
    \label{eq:specs}
\end{equation}
which capture the safety and progress requirements on the system, and the tester respectively.
We show that it is possible to model the set of feasible test executions using network flows on an automaton. 
% Traditionally, in computer science, network flows are studied in a single player setting, and the max-cut min-flow problem allows us to cut or constrain any of the edges. 
\begin{definition}[Network Flow] A \emph{network flow} is a tuple $\mathcal{N}=\langle V, E, c, s, t \rangle$ where $V$ is a set of vertices, $E$ is a set of directed edges, $E \subseteq V \times V$, $c$ is a capacity function for the amount of flow that each edge can transfer, and $s \in V$ are the source vertices and $t\in V$ are the target sink vertices. In a single network flow, each edge is associated with a single flow, while in a multi-commodity setting, multiple flows are associated with each edge. An extension of network flows to the game setting is known as a flow game~\cite{kupferman2018flow}. While our work involves solving a Stackleberg game over a multi-commodity flow network, it differs from~\cite{kupferman2018flow} in that the tester is not completely adversarial.
\end{definition}
% \begin{definition}[Flow Games~\cite{kupferman2018flow}]
% \label{def:flow-game}
% A flow game for two players denoted Player 0 and Player 1 is given as $\mathcal{G}=\langle V_0,V_1,E, c,s,t\rangle$ where $V_0 \cup V_1 \subseteq V$ where the set of vertices is partitioned between the two players. Player 0 directs the flow into each vertex in $V_0$ and its goal is to maximize the flow from $s$ to $t$, while Player 1 directs the flow into each vertex in $V_1$ and its goal is to minimize the flow from $s$ to $t$.
% \end{definition}

\section{Synthesizing Test Environments}
\vspace{-1mm}
% \textcolor{blue}{Make specific.}
This section sets up the test generation problem statement and introduces a running example to illustrate the approach we take in this paper.
\begin{figure*}
\centering
\vspace{2mm}
\begin{minipage}{.3\textwidth}
\begin{minipage}{\textwidth}
\centering
   \includegraphics[width=0.6\linewidth,trim={-1cm -1.2cm -1cm 0.0cm}]{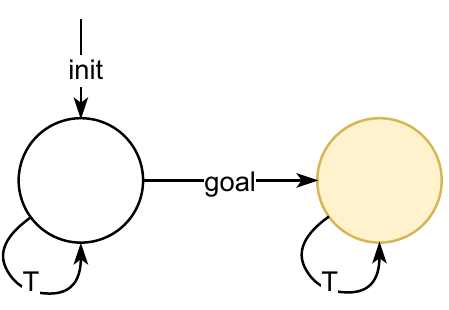}
   \vspace{-8mm}
   \subcaption{\label{fig:sys_ba} System Büchi automaton $B_{\text{sys}}$.}
  \end{minipage} 
  \begin{minipage}{\textwidth}
    \centering
    \vspace{2mm}
\includegraphics[width=0.6\linewidth,trim={0.0cm 0.0cm 0.0cm 0.0cm}]{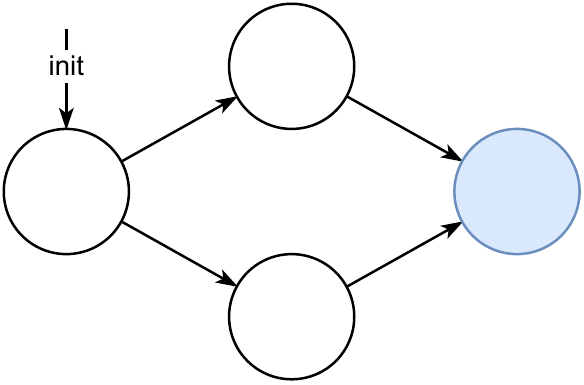}
    \subcaption{\label{fig:test_ba} Tester Büchi automaton $\mathcal{B}_{\text{test}}$.}
  \end{minipage}
 \end{minipage}
  \hspace{1mm}
  \begin{minipage}{.3\textwidth}
  \begin{minipage}{\textwidth}
      \centering
    \includegraphics[width=0.65\linewidth,trim={0.0cm 0.0cm 0.0cm 0.0cm}]{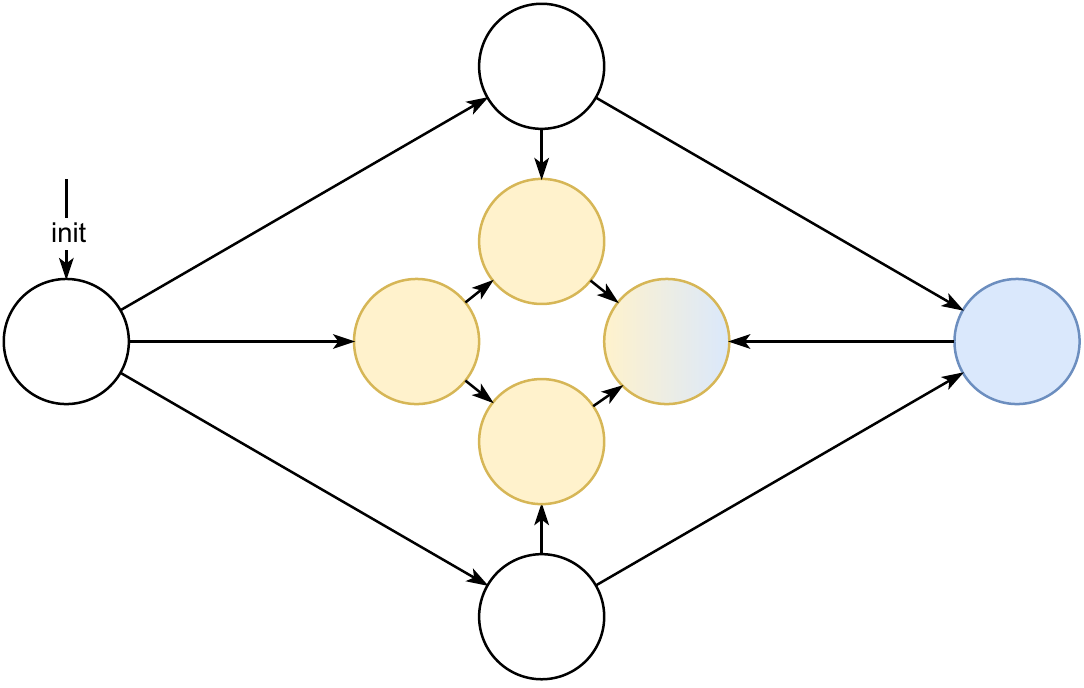}
    \vspace{-1mm}
    \subcaption{\label{fig:prod_ba} Specification product automaton $\mathcal{B}_\pi$.}
    \vspace{3mm}
  \end{minipage}
  \begin{minipage}{\textwidth}
  \centering
  \includegraphics[width=0.75\linewidth,trim={0.0cm 0.0cm 0.9cm 0.0cm}]{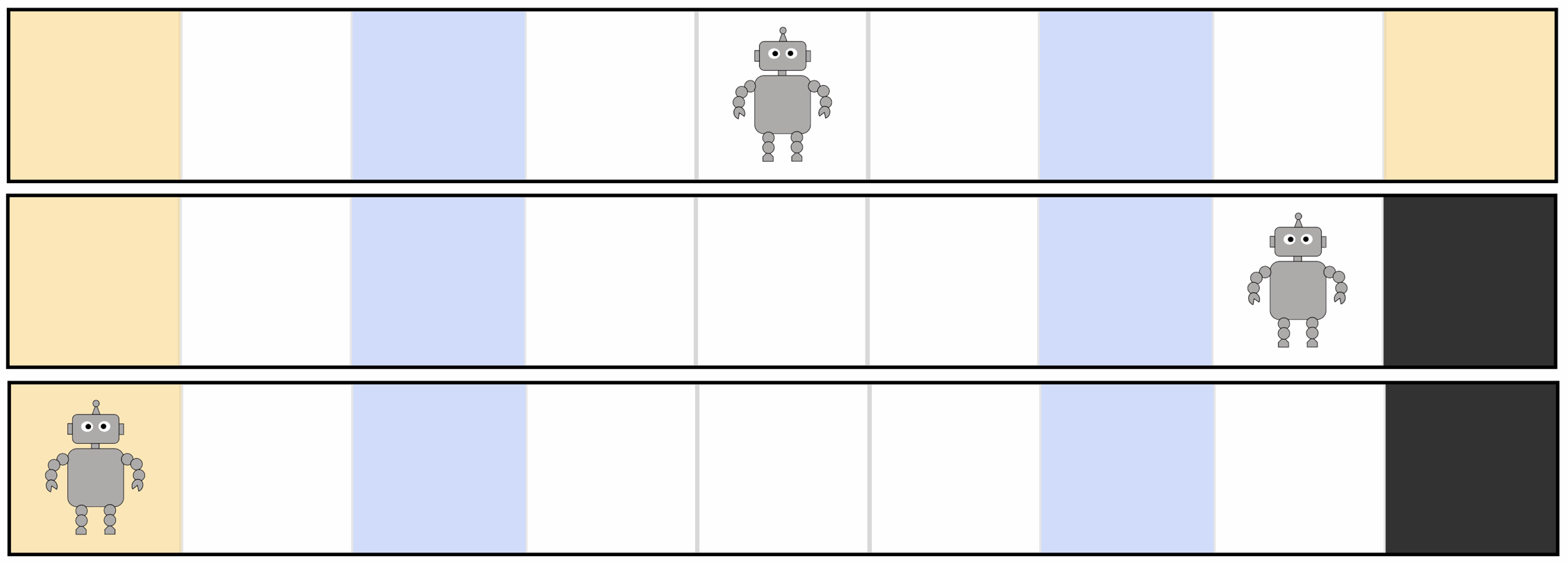}
  \subcaption{\label{fig:corridor_results} {Resulting Test Execution.}}
\end{minipage}  
\end{minipage}
  \hspace{2mm}
  \begin{minipage}{.3\textwidth}
  % \centering
  \centering
  \includegraphics[width=0.75\linewidth,trim={0.0cm 0.0cm 0.9cm 0.0cm}]{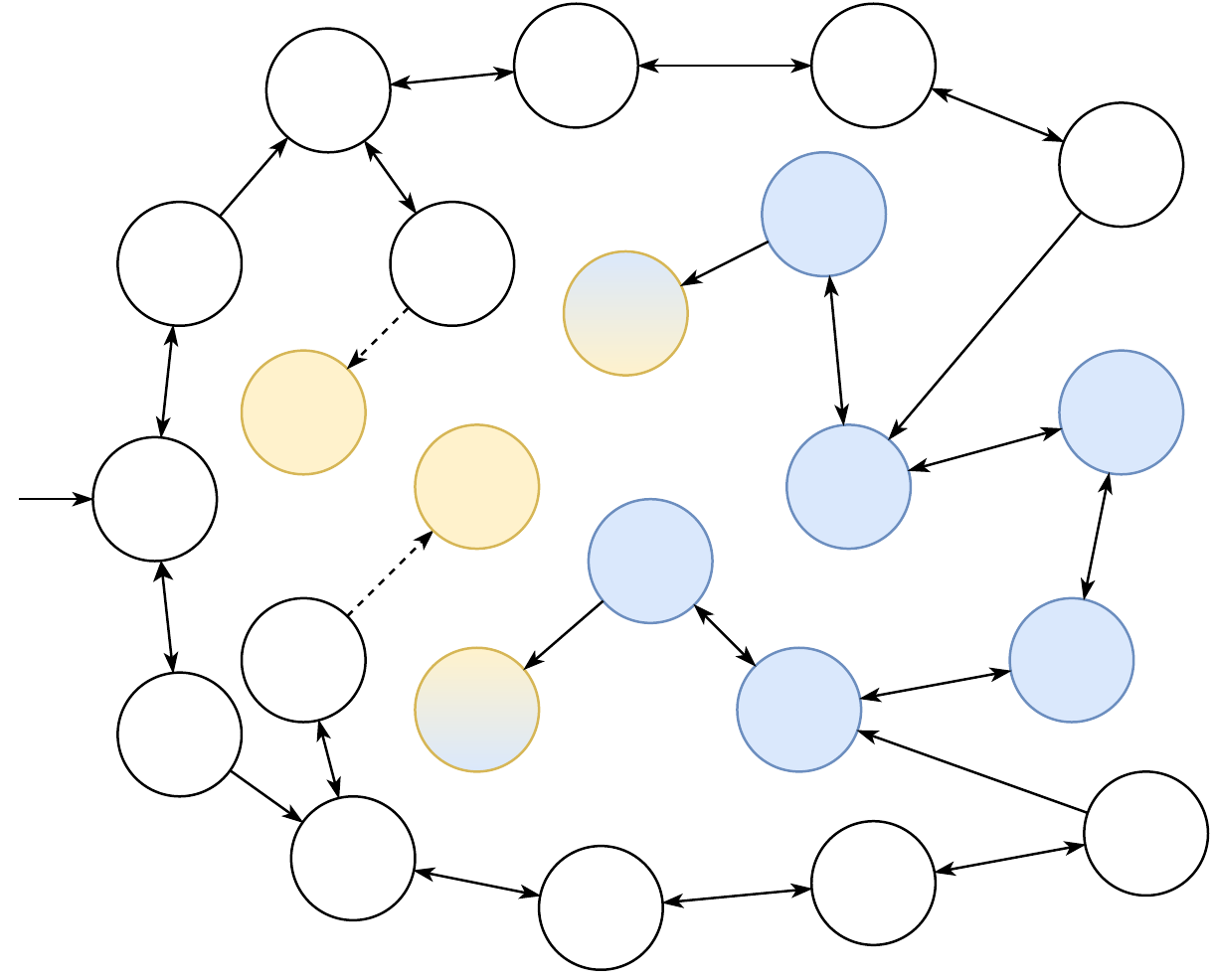}
  \subcaption{\label{fig:virtual} {Virtual game graph $\mc{G} = \mc{T} \otimes \mc{B}_\pi $. The initial state corresponds to $\mathtt{S}$, (partially) yellow states to $\mathtt{T}$, and exclusively blue states to $\mathtt{I}$.}}
  \end{minipage}

\caption{Büchi automata for the system specification, the test specification and the product automaton $\mc{B}_\pi = \mc{B}_{\text{test}} \times \mc{B}_{\text{sys}}$, and the virtual product graph $\mc{G}$ for the corridor navigation example. The accepting states of the system are shaded in yellow and the acceptance states of the tester are shaded in blue, with nodes shaded in both yellow and blue representing acceptance states of both tester and system. Transition labels and self-loops have been omitted in $\mc{B}_{\text{test}}$, $\mc{B}_\pi$, and $\mc{G}$ for clarity.}
\label{fig:ba_plot}
\vspace{-5mm}
\end{figure*}
\subsection{Problem Statement}
\label{sec:prob_statement}
\vspace{-1mm}
% The system and test specifications, \(\varphi_{\text{sys}}\) and \(\varphi_{\text{test}}\), represent, at a high-level abstraction, safety and progress requirements of the system and the test environment. 
The system and test specifications are written at the same level of abstraction as the model of the system characterized by the transition system \(\mathcal{T}\). We require that the sub-formulas of the test specifications, \(\varphi^s_{\text{test}}\) and \(\varphi^p_{\text{test}}\) in equation~\ref{eq:specs}, be high-level descriptions of desired test scenarios provided by a test engineer. In this way, the task of describing the behavior of the system to be observed during the test is left to the test engineer, but the process of synthesizing a corresponding test environment can be automated.
% \begin{remark}
% This framework of generating test scenarios is possible in use cases where high-level descriptions of desired test scenarios can be provided, for example in temporal logic. 
% In particular, we can address test generation for long-horizon planning tasks in which specifications can be characterized over discrete abstractions of the system. 
% The use cases in our paper will also illustrate how this framework is still relevant when applied to long-horizon tests on physical robots.
% \end{remark}
\begin{problem}
\label{prob1}
Given a discrete abstraction of a system model \(\mathcal{T} = (S, A, E, I, AP, L)\), and system and test specifications, \(\varphi_{\text{sys}}\) and \(\varphi_{\text{test}}\), defined over the set \(AP\), find the set of transitions of the system \(E_{\text{cut}} \subset E\) that need to be constrained such that all traces \(\sigma \models \varphi_{\text{sys}}\) of the constrained system on \(\mathcal{T} = (S, A, E_{\text{cut}}, I, AP, L)\) satisfy the following property: 
\vspace{-5mm}
\end{problem}
\begin{equation}\label{eq:prob}
    \sigma \models \varphi_{\text{sys}} \implies \sigma \models \varphi_{\text{test}}.
    \vspace{-1mm}
\end{equation}
In other words, a trace of the constrained system that satisfies the system specification must also satisfy the test specification, and the constraints are synthesized such that there always exists a trace satisfying the system specification.
In addition to determining constraints that result in test executions of the system abiding by equation~\eqref{eq:prob}, we do not want the system to be so constrained that it does not have much freedom in decision-making during the test. In this work, we use maximum network flow as a proxy for the maximum freedom a robot has to achieve its specifications, and we present an algorithmic framework that addresses both these problems on examples in both simulation and hardware. 
\subsection{Running Example: Robot in a corridor}
\label{sec:corridor}
% \begin{figure}
%     \centering
%     \includegraphics[width=\columnwidth]{figs/running_example_setup.png}
%     \caption{Running example, robot navigation in a corridor. The grid cells that the robot wants to reach are shaded in yellow; and the grid cells that correspond to the test specification are blue.}
%     \label{fig:corridor}
% \end{figure}
 Consider a corridor in a grid world shown in figure~\ref{fig:corridor_results}. The system under test is starting in the middle of the corridor with the goal of reaching either end. The dynamics are simple grid world dynamics enabling horizontal transitions to neighboring grid cells. The test behavior that we want to observe is that the system passes through the two blue grid cells $\varphi_{\text{test}}=\lozenge \text{key}_1 \land \lozenge \text{key}_2$. The system specification is given as $\varphi_{\text{sys}}= \lozenge \text{goal}$, which corresponds to the yellow grid cells.
We then constrain the system transitions according to our algorithm by placing obstacles on the grid cells, which we will show in the following sections.
\vspace{-1mm}
% \textcolor{blue}{Replace Pacman by corridor (?). Perhaps keep Pacman or road network for more intuition.}
% We will use the following as a running example throughout this paper. Consider a maze in a grid world setting. The system under test is an agent starting at its initial position on the bottom left corner of the grid with the intention of reaching its goal position in the top right corner. The dynamics are given as simple grid world dynamics enabling horizontal and vertical transitions to neighboring grid cells. The test behavior that we want to observe is that the system passes through specific grid cells, defined by the test specification. We then constrain the environment by placing obstacles on the grid cells by solving optimization~\ref{eq:min_node_cut}.
\subsection{Constructing Product Automata}
% \textcolor{blue}{Ask Richard if headline needs to be more specific(?). Product automata are well-known but here we introduce two special cases of this.}
\label{sec:prod_aut}
\vspace{-1mm}
We draw from automata theory to define the specification product automaton and virtual product graph. The product automaton of the system transition \(\mathcal{T}\) and the B\"uchi automaton corresponding to the specification \(\varphi_{\text{sys}}\), \(\mc{T}\otimes \mc{B}_{\text{sys}}\), is denoted by the tuple \(\mc{S} = (S_{\text{sys}}, A, E_{\text{sys}}, I_{\text{sys}}, AP_{\text{sys}}, L_{\text{sys}})\).
The asynchronous product is used to construct the specification product automaton of the system and test B\"uchi automata. The product automaton of the transition system, \(\mathcal{T}\), and the specification product automaton is denoted as the virtual product graph \(\mathcal{G}\). It is for this product automaton \(\mathcal{G}\), that we will be synthesizing constraints.

\begin{definition}[Specification Product Automaton]
The \emph{specification product automaton} \(\mc{B}_{\Pi} = \mc{B}_{\text{sys}} \times \mc{B}_{\text{test}}\) is the asynchronous product of the B\"uchi automata of the system and the test specification. 
In particular, $ \mc{B}_{\Pi}.F = \{(q_{\text{sys}}, q_{\text{test}}) \in \mc{B}_{\Pi}.Q | q_{\text{sys}} \in \mc{B}_{\text{sys}}.F\} \cup \{ (q_{\text{sys}}, q_{\text{test}}) \in \mc{B}_{\Pi}.Q |q_{\text{test}} \in \mc{B}_{\text{test}}.F \}$.
\end{definition}
\begin{definition}[Virtual Product Graph]
The synchronous product automaton of system transition \(\mathcal{T}\) and the NBA \(\mc{B}_{\Pi}\) is the \emph{virtual product graph} \(\mathcal{G} = \mathcal{T} \otimes \mc{B}_{\Pi}\). In tuple form, we denote \(\mc{G} = (S', A, E', I', AP', L')\).
\end{definition}
\begin{definition}[Source, Intermediate and Target Nodes]
The source ($\mathtt{S}$), intermediate ($\mathtt{I}$) and target ($\mathtt{T}$) are the set of nodes on the virtual product graph \(\mc{G}\) with the following properties:
\begin{equation}
    \begin{aligned}
        &\mathtt{S} = \{(s_0, q_0) \in S' \vert q_0 \in \mc{B}_{\Pi}.Q\} \\
        &\mathtt{I} = \{(s, (q_{\text{sys}}, q_{\text{test}})) \in S' \vert q_{\text{test}}\in \mc{B}_{\text{test}}.F,\, q_{\text{sys}}\notin \mc{B}_{\text{sys}}.F\} \\
       &\mathtt{T} = \{(s, (q_{\text{sys}}, q_{\text{test}})) \in S' \vert q_{\text{sys}}\in \mc{B}_{\text{sys}}.F\}\\
    \end{aligned}
\end{equation}

The source nodes \(\mathtt{S}\) represent the initial conditions of the test, the intermediate nodes \(\mathtt{I}\) represent the acceptance states corresponding to the test specification, and the target nodes \(\mathtt{T}\) represent the acceptance states corresponding to the system specification. For the running example, the automata corresponding to \(\mc{B}_{\text{sys}}\), \(\mc{B}_{\text{test}}\), \(\mc{B}_{\Pi}\) are illustrated in Figure~\ref{fig:sys_ba}-\ref{fig:prod_ba}. the virtual product graph \(\mc{G}\) and the corresponding source, intermediate and target nodes are illustrated in Figure~\ref{fig:virtual}.
\end{definition}
% For the robot navigation in a corridor example the Büchi automata encoding the test specifications are given in Figure~\ref{fig:sys_ba} and Figure~\ref{fig:test_ba}. We then construct the specification product automaton $\mc{B}_\pi$ and the virtual product graph $\mc{G}$, shown in Figure~\ref{fig:prod_ba} and in Figure~\ref{fig:virtual} respectively.
\begin{problem}
Given the setting in Problem~\ref{prob1}, synthesize the set of edge constraints \(E'_{\text{cut}}\) on the virtual product graph \(\mc{G}\) such that flows from \(\mathtt{S}\) to \(\mathtt{T}\) is maximized, and all possible traces on \(\mc{G}\) from \(\mathtt{S}\) to \(\mathtt{T}\) comprise of a node in \(\mathtt{I}\).
\label{prob2}
\end{problem}
\subsection{Multi-Commodity Flows and Bilevel Optimization}
\vspace{-1mm}
\label{sec:opt}
To synthesize constraints \(E'_{\text{cut}}\) on the virtual product graph \(\mc{G}\), we use multi-commodity flows on \(\mc{G}\) and present a bilevel optimization to find cuts. These constraints are such that a) there exists a system controller that can satisfy the system specification (\(\sigma \models \varphi_{\text{sys}}\)), and that for every satisfying trace, the test specification is also satisfied (equation~\eqref{eq:prob}), and b) the set of cuts \(E'_{\text{cut}}\) on \(\mc{G}\) result in maximum flow from \(\mathtt{S}\) to \(\mathtt{T}\). We then map these synthesized constraints \(E'_{\text{cut}}\) to constraints \(E_{\text{cut}}\) on system transitions \(\mathcal{T}\).

Given a graph \(\mc{G}\) with $\mathtt{S}$, $\mathtt{I}$, and $\mathtt{T}$, a brute force approach to solving Problems~\ref{prob1} and~\ref{prob2} is not viable, as it would involve a) finding a set of paths \(P_{\mathtt{S}\rightarrow \mathtt{I}}\) realizing max-flow from \(\mathtt{S}\) to \(\mathtt{I}\), and b) finding a set of paths \(P_{\mathtt{I}\rightarrow \mathtt{T}}\) realizing max flow from the \(\mathtt{I}\) to \(\mathtt{T}\), such that \(P_{\mathtt{S}\rightarrow \mathtt{I}}\) and \(P_{\mathtt{I}\rightarrow \mathtt{T}}\) are disjoint except for the intermediate \(\mathtt{I}\). Finding such a feasible pair of \(P_{\mathtt{S}\rightarrow \mathtt{I}}\) and \(P_{\mathtt{I}\rightarrow \mathtt{T}}\) would take exponential time because enumerating all paths is exponential in the size of the graph~\cite{cormen2009introduction}. 

To address this combinatorial problem, we formulate a bilevel optimization that relaxes edge cuts to take fractional values. These relaxed constraints, in addition to our choice of the objective function, makes the optimization tractable. While the cut values of some edges take on fractional values due to the relaxation, we find empirically that these fractional cuts are not relevant to constraining the flow. The system and tester are players that optimize for different flows on the same virtual product graph \(\mc{G}\). The system player maximizes flow \(f_{\mathtt{S}\rightarrow \mathtt{T}}\), defined from $\mathtt{S}$ to $\mathtt{T}$, with the flow into and out of the intermediate ($\mathtt{I}$) constrained to zero. These represent behaviors of the system satisfying \(\varphi_{\text{sys}}\) without satisfying \(\varphi_{\text{test}}\). The tester player: i) maximizes flow \(f_{\mathtt{S} \rightarrow \mathtt{I}}\) (defined from $\mathtt{S}$ to $\mathtt{I}$), ii) maximizes flow \(f_{\mathtt{I} \rightarrow \mathtt{T}}\) (defined from  $\mathtt{I}$ to $\mathtt{T}$), and iii) minimizes flow \(f_{\mathtt{S} \rightarrow \mathtt{T}}\) that bypasses the intermediate \(\mathtt{I}\). We use the multi-commodity network flows to simultaneously reason about several flows on \(\mc{G}\) --- maximizing flows \(f_{\mathtt{S} \rightarrow \mathtt{I}}\) and \(f_{\mathtt{I} \rightarrow \mathtt{T}}\), and cutting the flow \(f_{\mathtt{S} \rightarrow \mathtt{T}}\). However, unlike the canonical multi-commodity flow framework~\cite{vazirani2001approximation}, our flows do not compete for edge capacities. Instead, the flows are coupled by the placement of cuts by constraining flow along the edges. Therefore, the tester does not directly set the flow \(f_{\mathtt{S} \rightarrow \mathtt{T}}\), but indirectly constrains it by placing cuts on system transitions.
This multi-commodity flow-based bilevel optimization is given in~\eqref{eq:minmax}. The variables of the optimization are normalized by the total flow \(F\) on \(\mc{G}\), which is defined as follows,
\vspace{-2mm}
\begin{equation}
F = \min\{\sum_{v:(\mathtt{S},v) \in E'} f^{(\mathtt{S},v)}_{\mathtt{S} \rightarrow \mathtt{I}}, \sum_{v:(\mathtt{I},v) \in E'} f^{(\mathtt{I},v)}_{\mathtt{I} \rightarrow \mathtt{T}}\}.
\vspace{-2mm}
\label{eq:minflow}
\end{equation}
% \vspace{-2mm}
We require the auxiliary variable \(t := 1/F\) to re-write network flow constraints in the normalized form.
For every edge \(e\in E'\), let \(d^e\) represent the constraint on the edge --- \(d^e=t\) means that the edge \(e\) is cut or fully constrained and \(d^e = 0\) means that the edge \(e\) is unconstrained. Similarly, for every \(e \in E'\), \(f^e_{S \rightarrow \mathtt{I}}\), \(f^e_{\mathtt{I} \rightarrow \mathtt{T}}\), and \(f^e_{\mathtt{S} \rightarrow \mathtt{T}}\) are the respective flow values on edge \(e\) in \(\mc{G}\). As the outer (min) player, the tester variables are the flows \(f^e_{\mathtt{S} \rightarrow \mathtt{I}}\) and \(f^e_{\mathtt{I} \rightarrow \mathtt{T}}\), edge cuts \(d^e\), and the auxiliary variable \(t\). The objective function corresponds to the tester synthesizing constraints \(d^e\) such that total flow \(F\) is maximized while max-flow of \(f_{\mathtt{S}\rightarrow \mathtt{T}}\) is minimized. Likewise, the system player gets to maximize flow \(f_{\mathtt{S}\rightarrow \mathtt{T}}\). Next, the constraints of the bilevel optimization are detailed.
Capacity constraints determine the maximum flow allowed on an edge. The capacity constraints for normalized variables in this optimization are given as follows,
\vspace{-2mm}
\begin{equation}\tag{c1}
    \begin{aligned}
 \forall e\in E',  \quad & 0 \leq d^e \leq t, & 0 \leq f^e_{\mathtt{S}\rightarrow \mathtt{I}} \leq t, \\
 \text{ } & 0 \leq f^e_{\mathtt{S}\rightarrow \mathtt{T}} \leq t, & 0 \leq f^e_{\mathtt{I}\rightarrow \mathtt{T}} \leq t. \\
    \end{aligned}
    \label{eq:cap}
\end{equation}
Cut constraints correspond to the cut variable and flow variable of an edge competing for its capacity. For all \(k \in \{\mathtt{S}\rightarrow \mathtt{I}, \mathtt{I} \rightarrow \mathtt{T}, \mathtt{S} \rightarrow \mathtt{T}\}\), the cut constraints are as follows,
 \begin{equation}\tag{c2}
 \forall e\in E',  \quad  d^e + f^e_k \leq t.
 \label{eq:cut}
 \vspace{-1mm}
\end{equation}
Flow conservation constraints ensure that the total flow entering entering a node is equal to the total flow leaving the node (unless the node is a source or a target). For \(k \in \{\mathtt{S}\rightarrow \mathtt{I}, \mathtt{I} \rightarrow \mathtt{T}, \mathtt{S} \rightarrow \mathtt{T}\}\), the conservation constraints are as follows,
\vspace{-2mm}
\begin{equation}\tag{c3}
   \forall v \in S' \sum_{u: (u,v) \in E'} f^{(u,v)}_k = \sum_{u: (v,u) \in E'} f^{(v,u)}_k.
 \label{eq:cons}
\end{equation}
Finally, equation~\eqref{eq:minflow} can be re-written as the following constraint after normalizing the variables,
\begin{equation}\tag{c4}
    1 \leq \sum_{v: (\mathtt{S},v) \in E} f^{(\mathtt{S},v)}_{\mathtt{S}\rightarrow \mathtt{I}},\, \quad  1 \leq \sum_{v: (\mathtt{I},v) \in E} f^{(\mathtt{I},v)}_{\mathtt{I}\rightarrow \mathtt{T}}.
 \label{eq:cut}
\end{equation}

Our framework can synthesize test environments by constraining system actions, not by enforcing them. When the bilevel optimization finds constraints on the virtual product \(\mc{G}\), it takes into account tester behavior unknown to the system in the form of the test specification.  The framework guarantees that under the synthesized constraints, there always exists a system trace \(\sigma = s'_0s'_1\ldots\) on \(\mc{G}\) satisfying the system specification. However, this could return constraints for which at some temporal instances in the test execution, there is no path for satisfying trace on the system product automaton \(S\). During a test execution, these constraints on the system could potentially translate to the system, at that temporal instance, not being able to re-plan to satisfying its requirements. If possible, our framework should return constraints for which at every temporal instance of the test execution, the system should find a feasible path to satisfying its requirements. We add the following constraints to ensure that for the state \(\sigma_t = s'_t\) during the test, for which the state of the system corresponds to \(s_t\), there exists a path to the acceptance states of \(\varphi_{\text{sys}}\) on \(\mc{S}\). 

It is not necessary that all of these constraints on the virtual product graph \(\mc{G}\) be visible at any given system state \(s_t\). This feature allows for the tester to place constraints reactively to the system state \(s_t\). To reason about constraints that would be visible to the system at \(s_t\), we define mappings between the product automata. The first mapping, \(M_{B_{\Pi}\rightarrow \mc{G}}\), from the specification product automaton \(B_{\Pi}\) to \(\mc{G}\) defines the states of \(\mc{G}\) that are active for a state \(q \in B_{\Pi}.Q\):
\begin{equation}
M_{B_{\Pi}\rightarrow \mc{G}}(q) = \{(s, (q_{\text{sys}}, q_{\text{test}})) \in S'\vert q = (q_{\text{sys}}, q_{\text{test}})\}.
\end{equation}
At \(s'_t = (s_t, q_t)\) during the test execution, state \(q_t\) is active in the specification product automaton \(B_{\Pi}\). The outgoing edges of nodes in \(M_{B_{\Pi}\rightarrow \mc{G}}(q_t)\) represent the set of constraints that can be mapped to \(\mc{S}\) at \(q_t\). Let \(C_{\mc{G}}(q_t)\) denote these edges: 
\begin{equation}
    C_{\mc{G}}(q_t) = \{(e, d^e) \vert e = (u, v) \in E' \text{ s.t. } u\in M_{B_{\Pi}\rightarrow \mc{G}}(q_t)\}.
\end{equation}
By construction, every edge-constraint pair $(e,d_e)$ in \(C_{\mc{G}}(q_t)\) maps to an edge-constraint pair in \(\mc{S}\). Let \(V_{\mc{G}\rightarrow \mc{S}}\) map the nodes from \(\mc{G}\) to \(\mc{S}\). If \(v=(s,(q_{\text{sys}}, q_{\text{test}}) \in S'\), then \(V_{\mc{G}\rightarrow \mc{S}}(v) = (s, q_{\text{sys}}) \in S_{\text{sys}}\). Therefore, for every \(e=((u,v), d^e) \in C_{\mc{G}}(q_t)\), the corresponding edge-constraint pair on \(\mc{S}\) is  \(((V_{\mc{G}\rightarrow \mc{S}}(u),V_{\mc{G}\rightarrow \mc{S}}(v)), d^e)\). We denote this map as \(M_{\mc{G}\rightarrow \mc{S}}\), and the mapping is formally stated as follows,
\begin{equation}
    M_{\mc{G}\rightarrow \mc{S}}((u, v),d^e) = (((V_{\mc{G}\rightarrow \mc{S}}(u), V_{\mc{G}\rightarrow \mc{S}}(v)), d^e).
    \label{eq:mapping_G_S}
\end{equation}
Note that the constraint found on \(G\) is mapped one-to-one to the constraint on \(S\). This node mapping also provides the initial and acceptance states in \(S\) that are relevant denoted \(V_{\mc{G}\rightarrow \mc{S}}(\mathtt{S})\) and \(V_{\mc{G}\rightarrow \mc{S}}(\mathtt{T})\), respectively. 
Thus, the constraints that are active on \(\mc{S}\) at \(q \in \mc{B}_{\Pi}.Q\) are as follows,
\begin{equation}\tag{c5}
    C_{\mc{S}}(q_t) = \{M_{\mc{G}\rightarrow \mc{S}}(e,d^e)\vert (e, d^e) \in C_{\mc{G}}(q)\}.
    \label{eq:map_constraints}
\end{equation}
\begin{remark}
\(C_{\mc{S}}(q_t)\) represents the largest set of constraints that could be visible to the system at \(s_t\). Note that we say largest possible because the constraints visible to the system at \(s_t\) are the constraints on \(\mc{S}\) projected onto system transition \(\mc{T}\) at that instant. However, not all constraints in \(C_{\mc{S}}(q_t)\) might apply to the system, which is in state \(s_t\). 
\end{remark}
For every edge \(e \in E_{\text{sys}}\) and for every \(q \in \mc{B}_{\Pi}.Q\), let \(f^e_{\mc{S}}(q)\) denote the flow from source \(V_{\mc{G}\rightarrow \mc{S}}(\mathtt{S})\) to target \(V_{\mc{G}\rightarrow \mc{S}}(\mathtt{T})\). For brevity, we do not elaborate the constraints here, but the flow variables \(f^e_{\mc{S}}(q)\) must respect the standard network flow constraints in equations~\eqref{eq:cap}-\eqref{eq:cons}. We require the following condition to be satisfied:
\begin{equation}\tag{c6}
\begin{aligned}
    \forall q \in \mc{B}_{\Pi}.Q, \quad \sum_{e = (V_{\mc{G}\rightarrow \mc{S}}(\mathtt{S}), v ) \in E_{\text{sys}}}f^e_{\mc{S}}(q) \geq 1.
\end{aligned}
\label{eq:proj}
\end{equation}
Since the above constraint is defined from a fixed source \(V_{\mc{G}\rightarrow \mc{S}}(\mathtt{S})\) to target \(V_{\mc{G}\rightarrow \mc{S}}(\mathtt{T})\) on \(\mc{S}\), we assume that from every state \((s, q_{\text{sys}})\in S_{\text{sys}}\), there exists an edge to the source \(V_{\mc{G}\rightarrow \mc{S}}(\mathtt{S})\). For the examples of this paper with the system specifications of the class~\eqref{eq:specs}, this is always the case. In future work, we would like to prove these properties for a larger class of specifications and transition systems.
% ========================== CHECK THIS! ======================
% Furthermore, since we require check for constraint~\eqref{eq"proj} for flow from \(V_{\mc{G}\rightarrow \mc{S}}(S)\) to target \(V_{\mc{G}\rightarrow \mc{S}}(T)\), we require the following assumption to be true. 
% When does checking for flow from start to finish break?
Therefore, the bilevel optimization for synthesizing reactive constraints is as follows,
% Project the constrained edges from the virtual game graph onto the physical space by placing obstacles, obstacles will be lifted according to the state in the product automaton $\mc{B}_\pi$. 
% In addition we need to ensure that there will always be a path in $B_{\text{sys}} \otimes \mc{T}$ to the goal after placing the constraints to not create an impossible scenario. 
% On the game graph \(\mc{G}\), the goal of the system is to reach the goal states, while the test environment routes every test execution through the intermediate states.
% ============== if gradient descent works: ========================= %
% This problem can thus be framed as a constrained min-max optimization problem with dependent feasible sets, a convex-concave min-max Stackleberg game~\cite{goktas2021convex}. Our problem fits the optimization studied in~\cite{goktas2021convex} because of the dependent constraints. The objective function of this game is the flow-cut gap and the max flow for commodity-3 on $\bar{\mc{G}}$. The resulting min-max Stackleberg game can be solved by the gradient-descent methods introduced in~\cite{goktas2021convex}. This convex-concave min-max optimization can be formalized as follows, 
% \textcolor{red}{Modify variables and add conservation constraints}
% =============== if gradient descent works: ========================= %
\begin{equation}
\begin{aligned}
\text{MCF-OPT}(\lambda): & & \\
\argmin_{f^e_{\mathtt{S} \rightarrow \mathtt{I}} f^e_{\mathtt{I} \rightarrow \mathtt{T}}, d^e, t, f^e_{\mc{S}}(q)} \, & \argmax_{f^e_{\mathtt{S} \rightarrow \mathtt{T}}} \quad & t + \lambda\sum_{v:(s_3,v) \in E'} f^e_{\mathtt{S} \rightarrow \mathtt{T}} \\
&\textrm{s.t.} & \eqref{eq:cap}-\eqref{eq:proj},\\
   %& 1 \leq \sum_{v:(s_2,v) \in E} f_{I \rightarrow T},\\
   %   & \sum_{v:(u,v) \in E} f_{S \rightarrow I}_e = \sum_{v:(v,u) \in E} f_{S \rightarrow I}_e,\, v \notin \{s, t\}, \\
%   & \sum_{v:(u,v) \in E} f_{I \rightarrow T} = \sum_{v:(v,u) \in E} f_{I \rightarrow T},\, v \notin \{s, i\}, \\
%   & \sum_{v:(u,v) \in E} f_{S \rightarrow T}_e = \sum_{v:(v,u) \in E} f_{S \rightarrow T}_e,\, v \notin \{i, t\}, \\
\end{aligned}
\label{eq:minmax}
\end{equation}
where the regularization parameter \(\lambda\) penalizes the tester (and rewards the system) on \(f_{S\rightarrow T}\) flow. This optimization is in the form of a convex-concave min-max Stackleberg game with dependent constraint sets studied in\cite{goktas2021convex}, for which there always exists a solution.  

\begin{algorithm}
\caption{Constraining Virtual Product Graph \(\mc{G}\)}\label{alg1} \begin{algorithmic}[1] 
\Procedure{Automata}{$\mc{T}, \varphi_{\text{sys}}, \varphi_{\text{test}}$}
\State $\mc{B}_{\text{sys}} \leftarrow \text{BA}(\varphi_{\text{sys}})$ \Comment{System B\"uchi automaton}
\State $\mc{B}_{\text{test}} \leftarrow \text{BA}(\varphi_{\text{test}})$  \Comment{Tester B\"uchi automaton}
\State \(\mc{B}_{\Pi} \leftarrow \mc{B}_{\text{sys}} \times \mc{B}_{\text{test}}\) \Comment{Specification product}
\State \(\mc{S} \leftarrow  \mc{T} \otimes \mc{B}_{\text{sys}} \) \Comment{System product}
\State \(\mc{G} \leftarrow \mc{T} \otimes \mc{B}_{\Pi}\) \Comment{Virtual Product Graph}
\State \textbf{return} $\mc{G}, \mc{S}, \mc{B}_\pi, \mc{B}_{\text{sys}}, \mc{B}_{\text{test}}$
\EndProcedure
\State 
% ---------------------------------------------------------------------------- %
\Procedure{Constraints}{$\mc{T}, \varphi_{\text{sys}}, \varphi_{\text{test}}$}
\State $\mc{G}, \mc{S}, \mc{B}_\pi, \mc{B}_{\text{sys}}, \mc{B}_{\text{test}}$  $\gets\Call{Automata}{\mc{T}, \varphi_{\text{sys}}, \varphi_{\text{test}}}$
\State Identify nodes \(\mathtt{S}\), \(\mathtt{I}\), \(\mathtt{T}\) on \(\mc{G}\) 
\State Choose \(\lambda^*\) that maximizes \(F\) and cuts \(f_{S\rightarrow T}\) 
\State \(f^*_{\mathtt{S}\rightarrow \mathtt{I}}, f^*_{\mathtt{I}\rightarrow \mathtt{T}}, f^*_{\mathtt{S}\rightarrow \mathtt{T}}, d^*, t^* \leftarrow\) MCF-OPT\((\lambda^*)\)
\State \(E'_{\text{cuts}} = set()\) \Comment{To store cuts of \(\mc{G}\)}
\For \(\quad e \in E'\)
\If \(\quad d^{*,e} = 1\) \Comment{Ignore fractional cuts}
\State \(E'_{\text{cuts}} \leftarrow E'_{\text{cuts}} \cup e\) 
\EndIf
\EndFor
\State Verify \(\mc{G} = (S', A, E'\setminus E'_{\text{cuts}})\) has no \(f_{\mathtt{S} \rightarrow \mathtt{T}}\) flow.
\State \textbf{return} \(E'_{\text{cuts}}\)
\EndProcedure
\end{algorithmic}
\label{alg:find_virtual_game_graph}
\end{algorithm}
% where \(f^k_e \in \mathbb{R}^n \) represents the flow along edge \(e\in E\), and is defined for every \(e \in E\) and for each of the three commodities \(k = 1,2,3\). The variable \(d^e \in \mathbb{R}^n\) represents whether edge \(e\in E\) is cut (\(d^e=1\)) or not (\(d^e=0\)), and the auxiliary variable \(F \geq 0\) represents the maximum of the commodity-1 flow and commodity-2 flow. The regularization parameter \(\lambda\) places weight on commodity-3 flow. The \emph{max}-player is the system, which tries to maximize commodity-3 flow from \(S\) to \(T\), while the \emph{min}-player is the test environment, which optimizes to cut off commodity-3 flow while achieving a sparse cut of edges. The conservation and capacity constraints are not listed above but apply to each vertex and edge respectively. Note that a similar formulation could be extended to minimally constrained tests with node constraints.
% Remark: We assume that the system will only change its strategy if it is blocked by an obstacle.
% We find the cuts on $\mc{G}$ (to make sure that if there is a feasible solution on $\mc{G}$ we find it, while simultaneously ensuring that for every set of cuts mapped from $G$ to $T$(according to the state in $\mc{B}_\pi$ there exists a path in $\mc{S}$). Then we project the cuts to $\mc{B}_\pi$ and lift the obstacles according to the state in $\mc{B}_\pi$. Then we project to $\mc{S}$ 
\subsection{Projecting the constraints onto the physical space}
\vspace{-1mm}
\label{sec:proj}
The optimization returns cut edges $E'_{\text{cuts}}$ on the virtual product graph $\mc{G}$, which could be fractional values. Fully constrained edges are assigned $d^e$ values close to $t$, therefore we will only consider those edges to be cut. Lower fractional values for $d_e$ still allow flow to pass through and are not considered cut. We now need to map these cuts to the physical space to constrain the system's actions during the test execution. We define the projection 
\begin{equation}
    P_{\mc{G}\rightarrow \mc{T}} (g) = \{ s \in S| g = (s, (q_{\text{sys}},q_{\text{test}})\},
    \label{eq:proj_G_to_T}
\end{equation}
which maps each state $g$ in the virtual product graph $\mc{G}$ to its corresponding state in the transition system $\mc{T}$. This is a many-to-one mapping where multiple states in $\mc{G}$ will map to a single state in $\mc{T}$.
Additionally we define the projection 
\vspace{-3mm}
\begin{equation}
    P_{\mc{G}\rightarrow \mc{B}_\pi} (g) = \{ (q_{\text{sys}},q_{\text{test}}) \in \mc{B}_\pi.Q| g = (s, (q_{\text{sys}},q_{\text{test}}) \}
    \label{eq:proj_G_to_Bpi}
    \vspace{-2mm}
\end{equation} 
which maps each state $g$ in $\mc{G}$ to its corresponding state in $\mc{B}_\pi$. 
During the test execution we will keep track of the system state in $\mc{G}$, and when the system enters a state $g$ in $\mc{G}$ with an active cut, the corresponding transition from state $P_{\mc{G}\rightarrow \mc{T}}(g)$ in $\mc{T}$ will be constrained.
We use the projection defined in equation~\ref{eq:proj_G_to_Bpi} to determine a change of state in $\mc{B}_\pi$. For each state of the system in $\mc{B}_\pi$, the test environment will accumulate constraints. Upon transitioning to a state $g'$ that $P_{\mc{G}\rightarrow \mc{B}_\pi} (g')$ results in a change of state in $\mc{B}_\pi$, the obstacles that were placed previously will be removed and new obstacles will be placed according the active cuts on $\mc{G}$. This procedure is outlined in Algorithm~\ref{alg:reactive_test}.

This makes our framework reactive to the system state during the test execution, where finding static constraints on $\mc{G}$ results in a reactive test policy that constrains the system actions according to the observed behavior during the test.
\setlength{\textfloatsep}{7pt}% Remove \textfloatsep
\begin{algorithm}
\caption{Reactive Test Synthesis}\label{alg1} 
\begin{algorithmic}[1] 
\Procedure{Reactive Test}{$\mc{T}, \varphi_{\text{sys}}, \varphi_{\text{test}}$}
\State $E'_{\text{cuts}} \gets\Call{Constraints}{\mc{T}, \varphi_{\text{sys}}, \varphi_{\text{test}}}$
\State $g$ $\leftarrow$ $g_0 \in \mc{G}$
% \State $(q_{\text{sys}},q_{\text{test}})$ $\leftarrow$ $(q_{\text{sys}},q_{\text{test}})_0 \in \mc{B}_\pi$
\State $\mc{C} \leftarrow \emptyset$ \Comment{Initialize empty set of active cuts.}
\State $E_{\text{current}} \leftarrow e$ \Comment{Initially all transitions from $\mc{T}$.}
\While{\textbf{not} $q'_{\text{sys}} \in \mc{B}_{\text{sys}}.F$}
\State $g'$ $\leftarrow$ \text{update\_state$(g,\mc{G}, E_{\text{current}})$}
\State $(q'_{\text{sys}},q'_{\text{test}})$ $\leftarrow$ $P_{\mc{G}\rightarrow \mc{B}_\pi}(g')$ \Comment{Find state in $\mc{B}_\pi$.}
\If{$(q'_{\text{sys}},q'_{\text{test}}) \neq (q_{\text{sys}},q_{\text{test}})$}
\State $\mc{C} \leftarrow \emptyset$ \Comment{Remove all active cuts.}
\EndIf

\If{$\text{outgoing\_edge}_{\mc{G}}(g') \in (E'_{\text{cuts}})$} \Comment{Add cut.}
\State $\mc{C} \leftarrow \mc{C} \cup \text{outgoing\_edge}_{\mc{T}}(P_{\mc{G}\rightarrow \mc{T}}(g'))$
\EndIf
\State $E_{\text{current}} \leftarrow E \setminus \mc{C}$ \Comment{Update available transitions.}
\State $g \leftarrow g'$
\EndWhile
\EndProcedure
\end{algorithmic}
\label{alg:reactive_test}
\end{algorithm}
\begin{figure*}
\centering
 \vspace{2mm}
\begin{minipage}{.2\textwidth}
  % \hspace{3mm}

    \includegraphics[width=\linewidth,trim={0.0cm 0.0cm -0.0cm 0.0cm}]{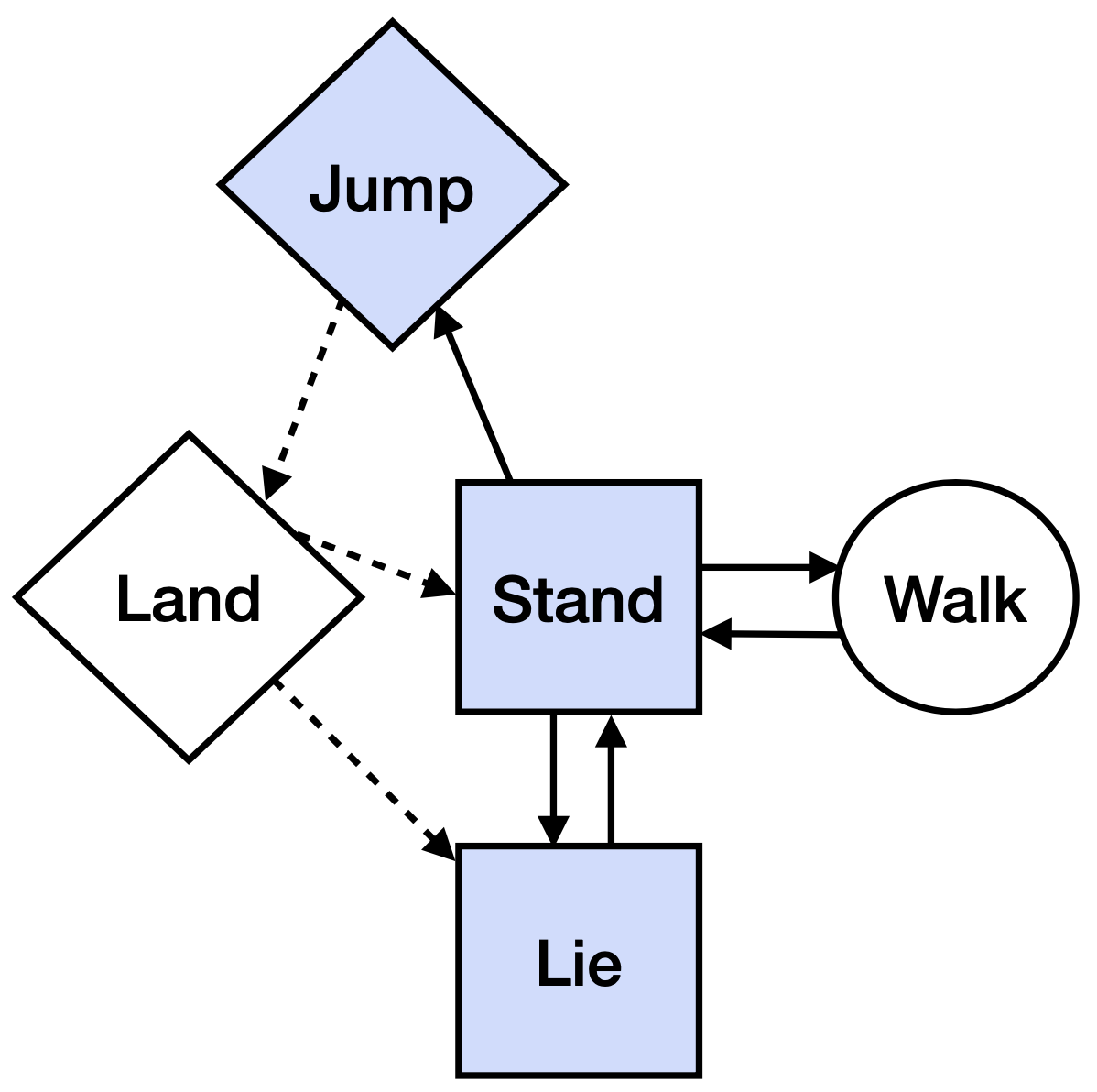}
    \subcaption{\label{fig:motionprims} Motion primitive graph.}
  \end{minipage}
  \hspace{3mm}
  \begin{minipage}{.65\textwidth}
    % \vspace{2mm}
    \centering
\includegraphics[width=\linewidth,trim={0.0cm 0.0cm 0cm 0.0cm}]{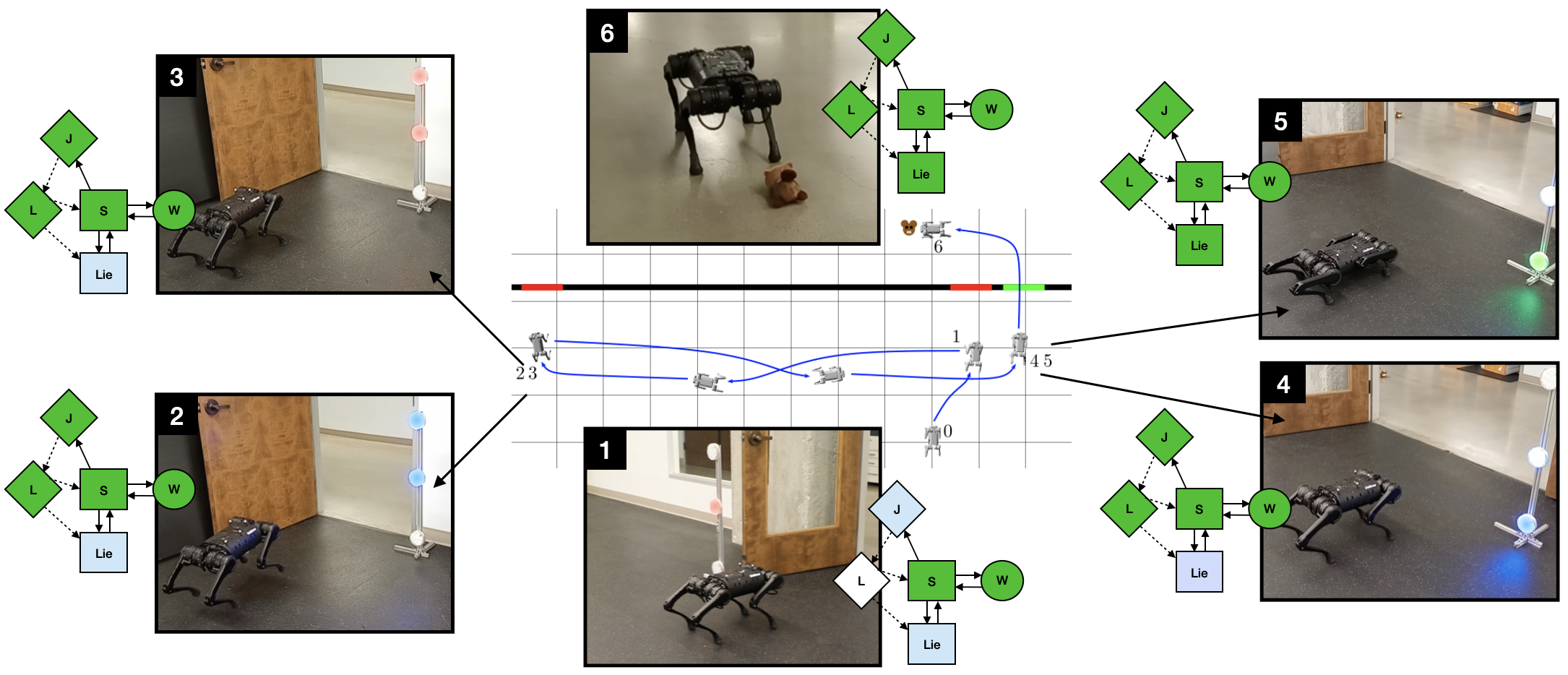}
    \subcaption{\label{fig:reactive_results} Snapshots of the hardware test execution on the Unitree A1 quadruped.}
  \end{minipage}
\caption{Resulting test execution on the Unitree A1 quadruped generated by this framework.}
\vspace{-4mm}
\label{fig:results}
\end{figure*}

% \begin{algorithm}
% \caption{Projecting constraints in $\mc{G}$ onto system \(\mc{T}\)}\label{alg1} \begin{algorithmic}[1] \Procedure{project}{$\mc{G}, \mc{B}_\pi, \mc{B}_{\text{sys}}, \mc{B}_{\text{test}}, \text{cuts}$}
% \State \text{to do}
% \EndProcedure
% \end{algorithmic}
% \label{alg:project}
% \end{algorithm}

% \begin{proposition}
% Algorithm~\ref{main} is sound.
% \end{proposition}
% \begin{proof}
% Need to prove that Algorithm~\ref{main}, if returning a non-empty set \(\rightarrow_{\text{cut}}\), will always return a solution to Problems~\ref{prob1} and \ref{prob2}.
% \end{proof}

% \begin{theorem}
% For any \(\mathcal{T}\), system and test specifications \(\varphi_{\text{sys}}\) and \(\varphi_{\text{test}}\), there exists a \(\lambda > 0\) such that a feasible solution to \end{theorem}
% \begin{proof}
% Need to prove that Algorithm~\ref{main}, if returning a non-empty set \(\rightarrow_{\text{cut}}\), will always return a solution to Problems~\ref{prob1} and \ref{prob2}.
% \end{proof}

% Projecting cuts onto physical space, defining the projection and proving that the flow does not change. We need to prove that the system controller always has a feasible path to the goal. Try proving this with where cuts are placed in the virtual product or introduce a mechanism to have this in place. 
 
% \begin{assumption} We assume that the system does not change its strategy until it encounters a constraint to its actions from the tester. 
% \end{assumption}
\section{Experimental Results}
\vspace{-1mm}
\label{sec:results}
We have implemented and validated this framework on simulated grid world examples and hardware experiments. In addition to these examples we have implemented the algorithm on grid world mazes and road networks: the results can be found in this GitHub repository\footnote{\href{https://github.com/abadithela/Flow-Constraints}{https://github.com/abadithela/Flow-Constraints}}.
% \begin{figure}
%     \centering
%     \includegraphics[width = \columnwidth]{figs/beaver_rescue.png}
%     \caption{Beaver Rescue test execution with the Unitree A1 quadruped. The robot exits the lab through the door and picks up the beaver, it then tries to enter the lab through the same door, which is now blocked. So the robot decides to walk down the hallway to enter the lab through the other door and successfully passes the test.}
%     \label{fig:beaver_rescue_results} 
%     \vspace{-5mm}
% \end{figure}
% \subsection{Simulated Examples}

\subsubsection{Robot in a Corridor}
\label{sec:robot_nav}
The agent under test for the running example is controlled by a simple grid world controller synthesized using TuLiP (Temporal Logic and Planning Toolbox)~\cite{wongpiromsarn2011tulip}.
The algorithm presented in section~\ref{sec:opt} results in a test execution during which the system under test visits the two pre-determined key locations before reaching one of the goal states at the end of the corridor. The resulting test execution is shown in Figure~\ref{fig:corridor_results}.

% \subsubsection{Search and Rescue: Three Door Example}\textcolor{blue}{Cutting this section out for space}
\subsubsection{Hardware Experiments with Quadruped}
Next we will find a test strategy to test an actual robotic system, the Unitree A1 quadruped.
This quadruped is controlled using a motion primitive layer with behaviors for lying down, standing, walking, and jumping. The underlying dynamics of the transitions between primitives are abstracted away from the higher-level autonomy as described in \cite{ubellacker2021iros}, and can be commanded directly.
%Individual motion primitives are implemented within our C++ motion primitive framework, and control laws, sensing, and estimation are executed at 1kHz on an Intel NUC with an i7-10710U CPU and 16GB of RAM. Communication to the A1's actuators and sensors is done via UDP.
The autonomy layer is provided by a TuLiP controller generated on an abstraction of the transition system of the quadruped, consisting of grid world locations and states corresponding to the available motion primitives.
We find test strategies and execute the resulting test for two test specifications inspired by a search and rescue mission.
\paragraph{Beaver Rescue}
The quadruped's task is to rescue the beaver from the hallway and return it to the lab, the system specification is given as $\varphi_{\text{sys}} = \lozenge \text{goal}$, where \textit{goal} corresponds to the quadruped and the beaver reaching the safe location in the lab.
The test specification is given as $\varphi_{\text{test}} = \lozenge \text{door}_1 \land \lozenge \text{door}_2$ ensuring that the quadruped will use a different door on the way to the beaver and back into the lab.
The resulting test execution shows the quadruped using \textit{door 2} to exit the lab into the hallway, after it reaches the beaver this door is shut and the quadruped walks to \textit{door 1} and returns the beaver to the safe location in the lab. 
Snapshots of this test execution can be seen in Figure~\ref{fig:overview}.
\paragraph{Search and Rescue: Motion Primitive Testing}
% \begin{figure}
%     \centering
%     \includegraphics[width=0.5\columnwidth]{figs/motion_primitive_graph.pdf}
%     \caption{Available motion primitives for the quadruped.}
%     \vspace{-5mm}
%     \label{fig:motionprims}
% \end{figure}
In this test we want to test the motion primitives of the quadruped shown in figure~\ref{fig:motionprims}. The goal for the quadruped is reaching the beaver in the hallway.
The test specification is given as $\varphi_{\text{test}} = \lozenge \text{jump} \land \lozenge \text{lie} \land \lozenge \text{stand}$, which ensures that we will test each of these motion primitive at least once.
The test setup includes lights at different heights, which correspond to the motion primitive which might unlock the door. The light starts in blue and after the motion primitive has been executed, the light will turn red - if the door remains locked - or green - if the door is unlocked. Our framework will decide whether the doors will be locked or unlocked according to which motion primitives have already been observed during the test. Snapshots of the test execution are shown in Figure~\ref{fig:reactive_results}.
\vspace{-1.2mm}
\section{Conclusions and Future Work}
\label{sec:Conclusions}
\vspace{-1.2mm}
We outlined the problem of finding the minimally constrained test as a bilevel optimization. For future work, we would like to prove that our algorithm is sound and complete, and provide sub-optimality guarantees on the generated test environment. The approach outlined in this paper requires reactive placement of obstacles during a test, which can be challenging for real-world use cases. Therefore, we aim to extend this framework to include dynamic test agents to constrain the system actions, and also allow for finding test environments requiring the minimum number of test agents to constrain a test environment for a test specification.

% \addtolength{\textheight}{-12cm}   % This command serves to balance the column lengths
                                  % on the last page of the document manually. It shortens
                                  % the textheight of the last page by a suitable amount.
                                  % This command does not take effect until the next page
                                  % so it should come on the page before the last. Make
                                  % sure that you do not shorten the textheight too much.

%%%%%%%%%%%%%%%%%%%%%%%%%%%%%%%%%%%%%%%%%%%%%%%%%%%%%%%%%%%%%%%%%%%%%%%%%%%%%%%%

%%%%%%%%%%%%%%%%%%%%%%%%%%%%%%%%%%%%%%%%%%%%%%%%%%%%%%%%%%%%%%%%%%%%%%%%%%%%%%%%

%%%%%%%%%%%%%%%%%%%%%%%%%%%%%%%%%%%%%%%%%%%%%%%%%%%%%%%%%%%%%%%%%%%%%%%%%%%%%%%%
% \section*{APPENDIX}
% Appendixes should appear before the acknowledgment.

\section*{ACKNOWLEDGMENTS}
The authors would like to acknowledge Mani Chandy, Tichakorn Wongpiromsarn, Qiming Zhao, Michel Ingham, Joel Burdick, Leonard Schulman, Shih-Hao Tseng, Ioannis Filippidis, and Ugo Rosolia for insightful discussions. 
\clearpage
%%%%%%%%%%%%%%%%%%%%%%%%%%%%%%%%%%%%%%%%%%%%%%%%%%%%%%%%%%%%%%%%%%%%%%%%%%%%%%%%
\balance
\bibliographystyle{IEEEtran}
\bibliography{refs}

% Generated by IEEEtran.bst, version: 1.14 (2015/08/26)
 \newcommand{\noop}[1]{}
\begin{thebibliography}{10}
\providecommand{\url}[1]{#1}
\csname url@samestyle\endcsname
\providecommand{\newblock}{\relax}
\providecommand{\bibinfo}[2]{#2}
\providecommand{\BIBentrySTDinterwordspacing}{\spaceskip=0pt\relax}
\providecommand{\BIBentryALTinterwordstretchfactor}{4}
\providecommand{\BIBentryALTinterwordspacing}{\spaceskip=\fontdimen2\font plus
\BIBentryALTinterwordstretchfactor\fontdimen3\font minus
  \fontdimen4\font\relax}
\providecommand{\BIBforeignlanguage}[2]{{%
\expandafter\ifx\csname l@#1\endcsname\relax
\typeout{** WARNING: IEEEtran.bst: No hyphenation pattern has been}%
\typeout{** loaded for the language `#1'. Using the pattern for}%
\typeout{** the default language instead.}%
\else
\language=\csname l@#1\endcsname
\fi
#2}}
\providecommand{\BIBdecl}{\relax}
\BIBdecl

\bibitem{sankaranarayanan2012falsification}
S.~Sankaranarayanan and G.~Fainekos, ``Falsification of temporal properties of
  hybrid systems using the cross-entropy method,'' in \emph{Proceedings of the
  15th ACM international conference on Hybrid Systems: Computation and
  Control}, 2012, pp. 125--134.

\bibitem{kapinski2016simulation}
J.~Kapinski, J.~V. Deshmukh, X.~Jin, H.~Ito, and K.~Butts, ``Simulation-based
  approaches for verification of embedded control systems: An overview of
  traditional and advanced modeling, testing, and verification techniques,''
  \emph{IEEE Control Systems Magazine}, vol.~36, no.~6, pp. 45--64, 2016.

\bibitem{annpureddy2011s}
Y.~Annpureddy, C.~Liu, G.~Fainekos, and S.~Sankaranarayanan, ``S-taliro: A tool
  for temporal logic falsification for hybrid systems,'' in \emph{International
  Conference on Tools and Algorithms for the Construction and Analysis of
  Systems}.\hskip 1em plus 0.5em minus 0.4em\relax Springer, 2011, pp.
  254--257.

\bibitem{chou2018using}
G.~Chou, Y.~E. Sahin, L.~Yang, K.~J. Rutledge, P.~Nilsson, and N.~Ozay, ``Using
  control synthesis to generate corner cases: A case study on autonomous
  driving,'' \emph{IEEE Transactions on Computer-Aided Design of Integrated
  Circuits and Systems}, vol.~37, no.~11, pp. 2906--2917, 2018.

\bibitem{dang2009coverage}
T.~Dang and T.~Nahhal, ``Coverage-guided test generation for continuous and
  hybrid systems,'' \emph{Formal Methods in System Design}, vol.~34, no.~2, pp.
  183--213, 2009.

\bibitem{hekmatnejad2020search}
M.~Hekmatnejad, B.~Hoxha, and G.~Fainekos, ``Search-based test-case generation
  by monitoring responsibility safety rules,'' \emph{arXiv preprint
  arXiv:2005.00326}, 2020.

\bibitem{plaku2013falsification}
E.~Plaku, L.~E. Kavraki, and M.~Y. Vardi, ``Falsification of ltl safety
  properties in hybrid systems,'' \emph{International Journal on Software Tools
  for Technology Transfer}, vol.~15, no.~4, pp. 305--320, 2013.

\bibitem{donze2010breach}
A.~Donz{\'e}, ``Breach, a toolbox for verification and parameter synthesis of
  hybrid systems,'' in \emph{International Conference on Computer Aided
  Verification}.\hskip 1em plus 0.5em minus 0.4em\relax Springer, 2010, pp.
  167--170.

\bibitem{fainekos2009robustness}
G.~E. Fainekos and G.~J. Pappas, ``Robustness of temporal logic specifications
  for continuous-time signals,'' \emph{Theoretical Computer Science}, vol. 410,
  no.~42, pp. 4262--4291, 2009.

\bibitem{dreossi2019verifai}
T.~Dreossi, D.~J. Fremont, S.~Ghosh, E.~Kim, H.~Ravanbakhsh,
  M.~Vazquez-Chanlatte, and S.~A. Seshia, ``Verifai: A toolkit for the formal
  design and analysis of artificial intelligence-based systems,'' in
  \emph{International Conference on Computer Aided Verification}.\hskip 1em
  plus 0.5em minus 0.4em\relax Springer, 2019, pp. 432--442.

\bibitem{DARPA_Urban_rules}
``{Technical Evaluation Criteria},''
  \url{https://archive.darpa.mil/grandchallenge/rules.html}.

\bibitem{DARPA_Urban}
``{DARPA Urban Challenge},''
  \url{https://www.darpa.mil/about-us/timeline/darpa-urban-challenge}.

\bibitem{kress2009temporal}
H.~Kress-Gazit, G.~E. Fainekos, and G.~J. Pappas, ``Temporal-logic-based
  reactive mission and motion planning,'' \emph{IEEE transactions on robotics},
  vol.~25, no.~6, pp. 1370--1381, 2009.

\bibitem{wongpiromsarn2012receding}
T.~Wongpiromsarn, U.~Topcu, and R.~M. Murray, ``Receding horizon temporal logic
  planning,'' \emph{IEEE Transactions on Automatic Control}, vol.~57, no.~11,
  pp. 2817--2830, 2012.

\bibitem{kloetzer2007temporal}
M.~Kloetzer and C.~Belta, ``Temporal logic planning and control of robotic
  swarms by hierarchical abstractions,'' \emph{IEEE Transactions on Robotics},
  vol.~23, no.~2, pp. 320--330, 2007.

\bibitem{lahijanian2015time}
M.~Lahijanian, S.~Almagor, D.~Fried, L.~E. Kavraki, and M.~Y. Vardi, ``This
  time the robot settles for a cost: A quantitative approach to temporal logic
  planning with partial satisfaction,'' in \emph{Twenty-Ninth AAAI Conference
  on Artificial Intelligence}, 2015.

\bibitem{censi2019liability}
A.~Censi, K.~Slutsky, T.~Wongpiromsarn, D.~Yershov, S.~Pendleton, J.~Fu, and
  E.~Frazzoli, ``Liability, ethics, and culture-aware behavior specification
  using rulebooks,'' in \emph{2019 International Conference on Robotics and
  Automation (ICRA)}.\hskip 1em plus 0.5em minus 0.4em\relax IEEE, 2019, pp.
  8536--8542.

\bibitem{shalev2017formal}
S.~Shalev-Shwartz, S.~Shammah, and A.~Shashua, ``On a formal model of safe and
  scalable self-driving cars,'' \emph{arXiv preprint arXiv:1708.06374}, 2017.

\bibitem{wongpiromsarn2021minimum}
T.~Wongpiromsarn, K.~Slutsky, E.~Frazzoli, and U.~Topcu, ``Minimum-violation
  planning for autonomous systems: Theoretical and practical considerations,''
  in \emph{2021 American Control Conference (ACC)}.\hskip 1em plus 0.5em minus
  0.4em\relax IEEE, 2021, pp. 4866--4872.

\bibitem{ernst2020arch}
G.~Ernst, P.~Arcaini, I.~Bennani, A.~Donze, G.~Fainekos, G.~Frehse,
  L.~Mathesen, C.~Menghi, G.~Pedrinelli, M.~Pouzet \emph{et~al.}, ``Arch-comp
  2020 category report: Falsification,'' \emph{EPiC Series in Computing}, 2020.

\bibitem{ghosh2018verifying}
S.~Ghosh, F.~Berkenkamp, G.~Ranade, S.~Qadeer, and A.~Kapoor, ``Verifying
  controllers against adversarial examples with bayesian optimization,'' in
  \emph{2018 IEEE International Conference on Robotics and Automation
  (ICRA)}.\hskip 1em plus 0.5em minus 0.4em\relax IEEE, 2018, pp. 7306--7313.

\bibitem{zutshi2014multiple}
A.~Zutshi, J.~V. Deshmukh, S.~Sankaranarayanan, and J.~Kapinski, ``Multiple
  shooting, cegar-based falsification for hybrid systems,'' in
  \emph{Proceedings of the 14th International Conference on Embedded Software},
  2014, pp. 1--10.

\bibitem{corso2021survey}
A.~Corso, R.~Moss, M.~Koren, R.~Lee, and M.~Kochenderfer, ``A survey of
  algorithms for black-box safety validation of cyber-physical systems,''
  \emph{Journal of Artificial Intelligence Research}, vol.~72, pp. 377--428,
  2021.

\bibitem{graebener2022towards}
J.~B. Graebener, A.~Badithela, and R.~M. Murray, ``Towards better test
  coverage: Merging unit tests for autonomous systems,'' in \emph{NASA Formal
  Methods: 14th International Symposium, NFM 2022, Pasadena, CA, USA, May
  24--27, 2022, Proceedings}, 2022, pp. 133--155.

\bibitem{apurva2022mincons}
A.~Badithela, J.~B. Graebener, and R.~M. Murray, ``Minimally constrained
  testing for autonomy with temporal logic specifications,'' 2022.

\bibitem{baier2008principles}
C.~Baier and J.-P. Katoen, \emph{Principles of model checking}.\hskip 1em plus
  0.5em minus 0.4em\relax MIT press, 2008.

\bibitem{Buchi1990}
\BIBentryALTinterwordspacing
J.~R. B{\"u}chi, \emph{On a Decision Method in Restricted Second Order
  Arithmetic}.\hskip 1em plus 0.5em minus 0.4em\relax New York, NY: Springer
  New York, 1990, pp. 425--435. [Online]. Available:
  \url{https://doi.org/10.1007/978-1-4613-8928-6_23}
\BIBentrySTDinterwordspacing

\bibitem{kupferman2018flow}
O.~Kupferman, G.~Vardi, and M.~Y. Vardi, ``Flow games,'' in \emph{37th IARCS
  Annual Conference on Foundations of Software Technology and Theoretical
  Computer Science (FSTTCS 2017)}.\hskip 1em plus 0.5em minus 0.4em\relax
  Schloss Dagstuhl-Leibniz-Zentrum fuer Informatik, 2018.

\bibitem{cormen2009introduction}
T.~H. Cormen, C.~E. Leiserson, R.~L. Rivest, and C.~Stein, \emph{Introduction
  to algorithms}.\hskip 1em plus 0.5em minus 0.4em\relax MIT press, 2009.

\bibitem{vazirani2001approximation}
V.~V. Vazirani, \emph{Approximation algorithms}.\hskip 1em plus 0.5em minus
  0.4em\relax Springer, 2001, vol.~1.

\bibitem{goktas2021convex}
D.~Goktas and A.~Greenwald, ``Convex-concave min-max stackelberg games,''
  \emph{Advances in Neural Information Processing Systems}, vol.~34, 2021.

\bibitem{wongpiromsarn2011tulip}
T.~Wongpiromsarn, U.~Topcu, N.~Ozay, H.~Xu, and R.~M. Murray, ``Tulip: a
  software toolbox for receding horizon temporal logic planning,'' in
  \emph{Proceedings of the 14th international conference on Hybrid systems:
  computation and control}, 2011, pp. 313--314.

\bibitem{ubellacker2021iros}
W.~Ubellacker, N.~Csomay-Shanklin, T.~G. Molnar, and A.~D. Ames, ``Verifying
  safe transitions between dynamic motion primitives on legged robots,'' in
  \emph{2021 IEEE/RSJ International Conference on Intelligent Robots and
  Systems (IROS)}, 2021, pp. 8477--8484.

\end{thebibliography}

\end{document}